# Unsupervised Joint Alignment and Clustering using Bayesian Nonparametrics

**Marwan A. Mattar**  **Allen R. Hanson**  **Erik G. Learned-Miller**

Department of Computer Science
University of Massachusetts Amherst, USA
http://vis-www.cs.umass.edu/

## Abstract

Joint alignment of a collection of functions is the process of independently transforming the functions so that they appear more similar to each other. Typically, such unsupervised alignment algorithms fail when presented with complex data sets arising from multiple modalities or make restrictive assumptions about the form of the functions or transformations, limiting their generality. We present a transformed Bayesian infinite mixture model that can simultaneously align and cluster a data set. Our model and associated learning scheme offer two key advantages: the optimal number of clusters is determined in a data-driven fashion through the use of a Dirichlet process prior, and it can accommodate any transformation function parameterized by a continuous parameter vector. As a result, it is applicable to a wide range of data types, and transformation functions. We present positive results on synthetic two-dimensional data, on a set of one-dimensional curves, and on various image data sets, showing large improvements over previous work. We discuss several variations of the model and conclude with directions for future work.

## 1 Introduction

Joint alignment is the process in which data points are transformed to appear more similar to each other, based on a criterion of joint similarity. The purpose of alignment is typically to remove unwanted variability in a data set, by allowing the transformations that reduce that variability. This process is widely applicable in a variety of domains. For example, removing temporal variability in event related potentials allows psychologists to better localize brain responses [32], removing bias in magnetic resonance images [21] provides doctors with cleaner images for their analyses, and removing (affine) spatial variability in images of objects can improve the performance of joint compression [9] and recognition [15] algorithms. Specifically, it has been found that using an aligned version of the Labeled Faces in the Wild [16] data set significantly increases recognition performance [5], even for algorithms that explicitly handle misalignments. Aside from bringing data into correspondence, the process of alignment can be used for other scenarios. For example, if the data are similar up to known transformations, joint alignment can remove this variability and, in the process, recover the underlying latent data [23]. Also, the resulting transformations from alignment have been used to build classifiers using a single training example [21] and learn sprites in videos [17].

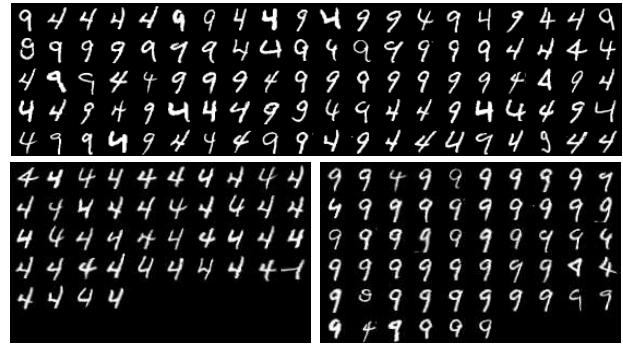

Figure 1: Joint alignment and clustering: given 100 unlabeled images (top), without any other information, our algorithm (§ 3) chooses to represent the data with two clusters, *aligns* the images and *clusters* them as shown (bottom). Our clustering accuracy is $94\%$, compared to $54\%$ with K-means using two clusters (using the minimum error across 200 random restarts). Our model is not limited to affine transformations or images.

### 1.1 Previous Work

Typically what distinguishes joint alignment algorithms are the assumptions they make about the data to be aligned and the transformations they can incur along with the level of supervision needed. Supervision takes several forms and can range from manually selecting landmarks to be aligned [5] to providing examples of data transformations [28]. In this paper we focus on unsupervised joint alignment which is helpful in scenarios where supervision is not practical or available. Several such algorithms exist.

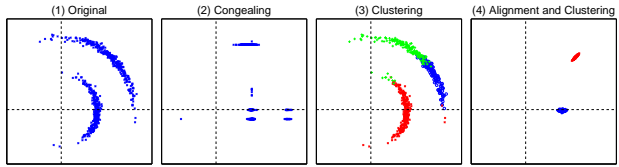

Figure 2: Illustrative example. (1) shows a data set of 2D points. The set of allowable transformations is rotations around the origin. (2) shows the result of the congealing algorithm which transforms points to minimize the sum of the marginal entropies. This independence assumption in the entropy computation causes the points to be squeezed into axis aligned groups. (3) highlights that clustering alone with an infinite mixture model may result in a larger number of clusters. (4) shows the result of the model presented in this paper. It discovers two clusters and aligns the points in each cluster correctly. This result is very close to the ideal one, which would have created tighter clusters.

In the curve domain, the *continuous profile model* [23] uses a variant of the hidden Markov model to locally transform each observation, while a mixture of regression model appended with global scale and translation transformations can simultaneously align and cluster [12]. Mattar *et al*. [25] adapted the *congealing* framework [21] to one dimensional curves. Congealing is an alignment framework that makes few assumptions about the data and allows the use of continuous transformations. It is a gradient-descent optimization procedure that searches for the transformations parameters that maximize the probability of the data under a kernel density estimate. Maximizing the likelihood is achieved by minimizing the entropy of the transformed data. It was initially applied to binary images of digits, but has since also been extended to grayscale images of complex objects [15] and 3D brain volumes [33]. Additionally, several congealing variants [31, 30, 6] have been presented that can improve its performance on binary images of digits and simple grayscale images of faces. Also in the image domain, the transformed mixture of Gaussians [11] and the work of Lui *et al*. [24] are used to align and cluster.

One of the attractive properties of congealing is a clear separation between the transformation operator and optimization procedure. This has allowed congealing to be applied to a wide range of data types and transformation functions [1, 15, 21, 22, 25, 33]. Its main drawback is its inability to handle complex data sets that may contain multiple modes (i.e. images of the digits 1 and 7). While congealing's use of an entropy-based objective function can in theory allow it to align multiple modes, in practice the independence assumption (temporally for curves and spatially for images) can cause it to collapse modes (see Figure 2 for an illustration). Additionally, its method for regularizing parameters to avoid excessive transformations is *ad hoc* and does not prevent it from annihilating the data (shrinking to size zero) in some scenarios.

### 1.2 Our Approach

The problem we address here is joint alignment of a data set that may contain multiple groups or clusters. Previous nonparametric alignment algorithms (e.g. congealing [21]) typically fail to acknowledge the multi-modality of the data set resulting in poor performance on complex data sets. We address this by simultaneously aligning and clustering [11, 12, 24] the data set. As we will show (and illustrated in Figure 2), solving both alignment and clustering together offers many advantages over clustering the data set first and then aligning the points in each cluster.

To this end, we developed a nonparametric[1] Bayesian joint alignment and clustering model that is a generalization of the standard Bayesian infinite mixture model. Our model possesses many of the favorable characteristics of congealing, while overcoming its drawbacks. More specifically, it:

- Explicitly clusters the data which provides a mechanism for handling complex data sets. Furthermore, the use of a Dirichlet process prior enables learning the number of clusters in a data-driven fashion.

- Can use any generic transformation function parameterized by a vector. This decouples our model from the specific transformations which allows us to plug in different functions for different data types.

- Enables the encoding of prior beliefs regarding the degree of variability in the data set, as well as regularizes the transformation parameters in a principled way by treating them as random variables.

We first present a Bayesian joint alignment model (§ 2) that assumes a unimodal data set (i.e. only one cluster). This model is a special case of our proposed joint alignment and clustering model that we introduce in § 3. We then discuss several variations of our model in § 4 and conclude in § 5 with directions for future work.

### 1.3 Problem Definition

We are provided with a data set $\mathbf{x} = \{x_i\}_{i=1}^N$ of $N$ items and a transformation function, $x_i = \tau(y_i, \rho_i)$ parameter-ized by $\rho_i$. Our objective is to recover the set of transformation parameters $\{\rho_i\}_{i=1}^{N}$, such that the aligned data set $\{y_i = \tau(x_i, \rho_i^{-1})\}_{i=1}^{N}$ is more coherent. In the process, we also learn a clustering assignment $\{z_i\}_{i=1}^{N}$ of the data points. Here $\rho_i^{-1}$ is defined as the parameter vector generating the inverse of the transformation that would be generated by the parameter vector $\rho$ (i.e. $x_i = \tau(\tau(x_i, \rho_i^{-1}), \rho_i)$).

---

[1]Here, we use the term nonparametric to imply that the number of model parameters can grow (a property of the infinite mixtures), and not that the distributions are not parametric.

## 2 Bayesian Joint Alignment

The Bayesian alignment (BA) model assumes a unimodal data set (in § 3 this assumption is relaxed). Consequently there is a single set of parameters ($\theta$ and $\rho$) that generate the entire data set (see Figure 3). Under this model, every observed data item, $x_i$, is generated by transforming a canonical data item, $y_i$, with transformation, $\rho_i$. More formally, $x_i = \tau(y_i, \rho_i)$, where $y_i \sim F_D(\theta)$ and $\rho_i \sim F_T(\varphi)$. The auxiliary variable $y_i$ is not shown in the graphical model for simplicity. Given the Bayesian setting, the parameters $\theta$ and $\varphi$ are random variables, with their respective prior distributions, $H_D(\lambda)$ and $H_T(\alpha)$.[2]

The model does not assume that there exists a single perfect canonical example that explains all the data, but uses a parametric distribution $F_D(\theta)$ to generate a slightly different canonical example, $y_i$, for each data item, $x_i$. This enables it to explain variability in the data set that may not be captured with the transformation function alone. The model treats the transformation function as a black-box operation, making it applicable to a wide range of data types (e.g. curves, images, and 3D MRI scans), as long as an appropriate transformation function is specified.

For both this model and the full joint alignment and clustering model introduced in the next section we use exponential family distributions for $F_D(\theta)$ and $F_T(\varphi)$ and their respective conjugate priors for $H_T(\alpha)$ and $H_D(\lambda)$. This allows us to use Rao-Blackwellized sampling schemes [4] by analytically integrating out the model parameters and caching sufficient statistics for efficient likelihood computations. Furthermore, the hyperparameters now play intuitive roles where they act as a pseudo data set and are easier to set or learn from data.

### 2.1 Learning

Given a data set $\{x_i\}_{i=1}^{N}$ we wish to learn the parameters of this model ($\{\rho_i\}_{i=1}^{N}, \theta, \varphi$). We use a Rao-Blackwellized Gibbs sampler that integrates out the model parameters, $\theta$ and $\varphi$, and only samples the hidden variables, $\{\rho_i\}_{i=1}^{N}$. Such samplers typically speed-up convergence. The intuition is that the model parameters are implicitly updated with the sampling of every transformation parameter instead of once per Gibbs iteration. The resulting Gibbs sam-

---

[2]Here we assume that the hyperparameters $\alpha$ and $\lambda$ are fixed, but they can be learned or sampled if necessary.

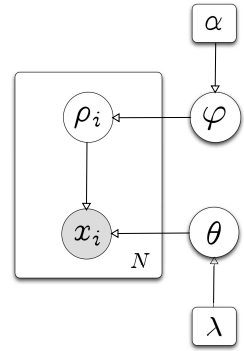

Figure 3: Graphical representation for our proposed Bayesian alignment model (§ 2).

pler only iterates over the transformation parameters:

$$\begin{aligned}
\forall_{i=1:N} \ \rho_i^{(t)} &\sim p(\rho_i|\mathbf{x}, \boldsymbol{\rho}_{-i}^{(t)}, \alpha, \lambda) \\
&\propto p(\rho_i, x_i|\mathbf{x}_{-i}, \boldsymbol{\rho}_{-i}^{(t)}, \alpha, \lambda) \\
&= p(x_i|\rho_i, \mathbf{x}_{-i}, \boldsymbol{\rho}_{-i}^{(t)}, \lambda) p(\rho_i|\boldsymbol{\rho}_{-i}^{(t)}, \alpha) \\
&= p(y_i|\mathbf{y}_{-i}^{(t)}, \lambda) p(\rho_i|\boldsymbol{\rho}_{-i}^{(t)}, \alpha),
\end{aligned}$$

where $y_i = \tau(x_i, \rho_i^{-1})$. The $t$ superscript in the above equations refers to the Gibbs iteration number.

Sampling $\rho_i$ is complicated by the fact that $p(y_i|\mathbf{y}_{-i}^{(t)}, \lambda)$ depends on the transformation function. Previous alignment research [21, 24], has shown that the gradient of an alignment objective function with respect to the transformations provides a strong indicator for how alignment should proceed. One option would be Hamiltonian Monte Carlo sampling [26] which uses the gradient as a drift factor to influence sampling. However, instead of relying on direct sampling techniques, we use approximations based on the posterior mode [13]. Such an approach is more direct since it is expected that the distribution will be tightly concentrated around the mode. Thus, at each iteration the transformation parameter is updated as follows:

$$\rho_i = \arg\max_{\rho_i} p(y_i|\mathbf{y}_{-i}^{(t)}, \lambda) p(\rho_i|\boldsymbol{\rho}_{-i}^{(t)}, \alpha).$$

Interestingly, the same learning scheme can be derived using the incremental variant [27] of hard-EM.

### 2.2 Model Characteristics

The objective function optimized in our model contains two key terms, a data term, $p(\mathbf{x}|\boldsymbol{\rho}, \theta)$, and a transformation term, $p(\boldsymbol{\rho}|\varphi)$. The latter acts as a regularizer to penalize large transformations and prevent the data from being annihilated. One advantage of our model is that large transformations are penalized in a principled fashion. More specifically, the cost of a transformation, $\rho_i$ is based on the

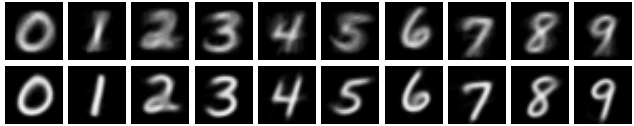

Figure 4: Top row. Means before alignment. Bottom row. Means after alignment with BA. The averages of pixelwise entropies are as follows. Before: 0.3 (top), with congealing: 0.23 (not shown), and with BA: 0.21 (bottom).

*learned* parameter $\varphi$ which depends on the transformations of all the other data items, $\boldsymbol{\rho}_{-i}$, and the hyperparameters, $\alpha$. Learning $\varphi$ from the data is a more effective means for assigning costs than handpicking them.

The model has several other favorable qualities. It is efficient, can operate on large data sets while maintaining a low memory footprint, allows continuous transformations, regularizes transformations in a principled way, is applicable to a large variety of data types, and its hyperparameters are intuitive to set. Its main drawback is the assumption of a unimodal data set, which we remedy in § 3. We first evaluate this model on digit and curve alignment.

### 2.3 Experiments

**Digits.** We selected 50 images of every digit from the MNIST data set and performed alignment on each digit class independently. The mean images before and after alignment are presented in Figure 4. We allowed 7 affine image transformations: scaling, shearing, rotating and translation. $F_D(\theta)$ is the product of independent Bernoulli distributions, one for each pixel location, and $F_T(\varphi)$ is a $7-$D zero mean diagonal Gaussian. For comparison, we also ran the congealing algorithm (see Figure 4).

**Curves.** We generated 85 curve data sets in a manner similar to curve congealing [25], where we took five original curves from the UCR repository [18] and for each one generated 17 data sets, each containing 50 random variations of the original curve. We used the same transformation function in curve congealing [25], which allows non-linear time warping (4 parameters), non-linear amplitude scaling (8 parameters), linear amplitude scaling (1 parameter), and amplitude translation (1 parameter). $F_D(\theta)$ was set to a diagonal Gaussian distribution (i.e. we treat the raw curves as a random vector), and $F_T(\varphi)$ was a $14-$D zero mean diagonal Gaussian. Again, we compared against the curve congealing algorithm. We computed a standard deviation score by summing the standard deviation at each time step of the final alignment produced by both algorithms. Figure 5 shows a scatter plot of these scores obtained by congealing and BA for all 85 data sets, as well as sample alignment results on two difficult cases. As the figure shows, the curve data sets can be quiet complex.

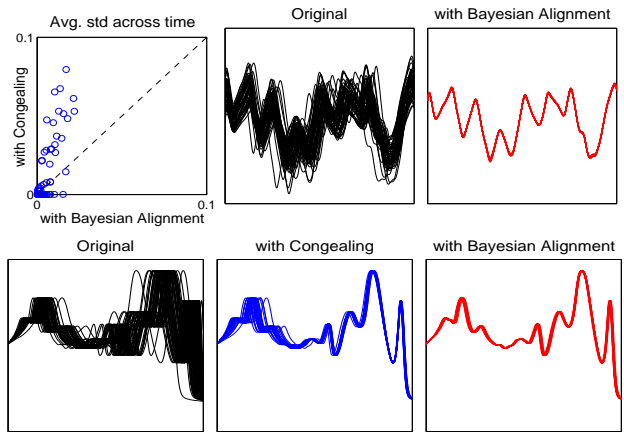

Figure 5: Top row. Left: scatter plot of the standard deviation score (see text) of congealing and the Bayesian alignment algorithm across the 85 synthetic curve data sets. Middle: An example of a difficult data set. Right: The alignment result of the difficult data set. The bottom row shows an example where BA outperformed congealing.

**Discussion.** On the digits data sets, BA performed at least as well as congealing for every digit class and on average performed better. On the curves data sets, BA does substantially better than congealing in many cases, but in some cases congealing does slightly better. In all the experiments, both congealing and BA converged. BA's advantage is largely due to its explicit regularization of transformations which enables it to perform a maximization at each iteration. Congealing's lack of such regularization requires it to take small steps at each iteration making it more susceptible to local optima. Furthermore, congealing typically requires five times the number of iterations to converge.

## 3 Clustering with Dirichlet Processes

We now extend the BA model introduced in the previous section to explicitly cluster the data points. This provides a mechanism for handling complex data sets that may contain multiple groups.

The major drawback of the BA model is that a single pair of data and transformation parameters ($\theta$ and $\varphi$, respectively) generate the entire data set. One natural extension to this generative process is to assume that we have several such parameter pairs (finite but unknown a priori) and each data point samples its parameter pair. By virtue of points sampling the same parameter pair, they are assigned to the same group or cluster. A Dirichlet process (DP) provides precisely this construction and serves as the prior for the data and transformation parameter pairs.

A DP essentially provides a distribution over distributions, or, more formally, a distribution on random probability measures. It is parameterized by a base measure and a concentration parameter. A draw from a DP generates a finite set of samples from the base measure (the concentration parameter controls the number of samples). A key advantage of DP's is that the number of unique parameters (i.e. clusters) can grow and adapt to each data set depending on its size and characteristics. Under this new probability model, data points are generated in the following way:

1. Sample from the DP, $G \sim DP(\gamma, H_\alpha \times H_\lambda)$. $\gamma$ is the concentration parameter, and $H_\alpha$ and $H_\lambda$ are the base measures for $F_T(\varphi)$ and $F_D(\theta)$ respectively.

2. For each data point, $x_i$, sample a data and transformation parameter pair, $(\theta_i, \varphi_i) \sim G$.

3. Sample a transformation and canonical data item from their distributions, $y_i \sim F_D(\theta_i)$ and $\rho_i \sim F_T(\varphi_i)$.

4. Transform the canonical data item to generate the observed sample, $x_i = \tau(y_i, \rho_i)$.

Figure 6 depicts the generative process as described above (distributional form, right) and in the more traditional graphical representation with the cluster random variable, $z$, and mixture weights, $\pi$, made explicit (left).

Our model can thus be seen as an extension of the standard Bayesian infinite mixture model where we introduced an additional latent variable, $\rho_i$, for each data point to represent its transformation. Several existing alignment models [11, 12, 21, 23] can be viewed as similar extensions to other standard generative models. Sometimes the transformations are applied to other model parameters instead of data points as in the case of transformed Dirichlet processes (TDP) [29]. TDP is an extension of *hierarchical Dirichlet processes* where global mixture components are transformed before being reused in each group. The challenge in introducing additional latent variables is in designing efficient learning schemes that can accommodate this increase in model complexity.

### 3.1 Learning

We consider two different learning schemes for this model. The **first** is a blocked, Rao-Blackwellized Gibbs sampler, where we sample both the cluster assignment $z_i$, and transformation parameters $\rho_i$, simultaneously:

$$(z_i^{(t)}, \rho_i^{(t)}) \sim p(z_i, \rho_i | \mathbf{z}_{-i}^{(t)}, \boldsymbol{\rho}_{-i}^{(t)}, \mathbf{x}, \gamma, \alpha, \lambda)$$
$$\propto p(z_i | \mathbf{z}_{-i}^{(t)}, \gamma) p(\rho_i | \boldsymbol{\rho}_{-i}^{(t)}, \alpha) p(y_i | \mathbf{y}_{-i}^{(t)}, \lambda).$$

As with the BA model, we approximate $p(\rho_i | \boldsymbol{\rho}_{-i}^{(t)}, \alpha) p(y_i | \mathbf{y}_{-i}^{(t)}, \lambda)$ with a point estimate based on its mode. Consequently this learning scheme is a direct generalization of the one derived for the BA model. Note

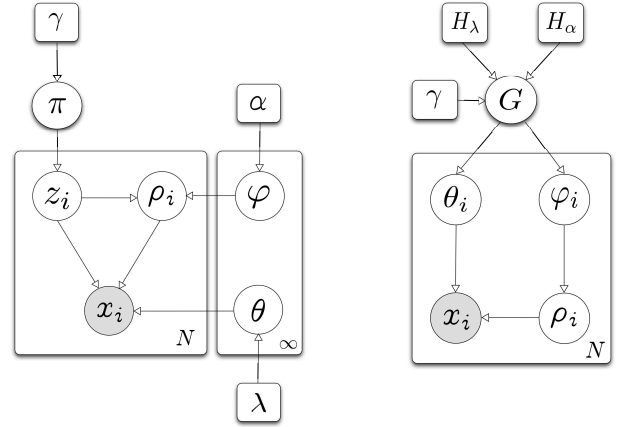

Figure 6: Graphical representation for our proposed non-parametric Bayesian joint alignment and clustering model (left) and its corresponding distributional form (right).

that $p(z_i | \mathbf{z}_{-i}^{(t)}, \gamma)$ is the cluster predictive distribution based on the Chinese restaurant process (CRP) [2].

While this sampler is effective (it produced the positive result in Figure 1) it scales linearly with the number of clusters and computing the most likely transformation for a cluster is an expensive operation. We designed an alternative sampling scheme that does not require the expensive mode computation and whose running time is independent of the number of clusters.

The **second** sampler further integrates out the transformation parameter, and only samples the cluster assignment. We now derive an implementation for this sampler.

$$\forall_{i=1:N} \ z_i^{(t)} \sim p(z_i | \mathbf{z}_{-i}^{(t)}, \mathbf{x}, \gamma, \alpha, \lambda)$$
$$\propto p(z_i, x_i | \mathbf{z}_{-i}^{(t)}, \mathbf{x}_{-i}, \gamma, \alpha, \lambda)$$
$$= p(z_i | \mathbf{z}_{-i}^{(t)}, \gamma) p(x_i | \mathbf{z}^{(t)}, \mathbf{x}_{-i}, \alpha, \lambda).$$

$p(x_i | \mathbf{z}, \mathbf{x}_{-i}, \alpha, \lambda)$

$$= \int_{\boldsymbol{\theta}} \int_{\boldsymbol{\varphi}} \int_{\rho_i} p(x_i, \rho_i, \boldsymbol{\theta}, \boldsymbol{\varphi} | \mathbf{z}, \mathbf{x}_{-i}, \alpha, \lambda) \ d\rho_i \ d\boldsymbol{\varphi} \ d\boldsymbol{\theta}$$

$$= \int_{\boldsymbol{\theta}} \int_{\boldsymbol{\varphi}} \left( \int_{\rho_i} p(x_i, \rho_i | z_i, \boldsymbol{\theta}, \boldsymbol{\varphi}, \alpha, \lambda) \ d\rho_i \right) \cdots$$
$$p(\boldsymbol{\theta}, \boldsymbol{\varphi} | \mathbf{z}_{-i}, \mathbf{x}_{-i}, \alpha, \lambda) \ d\boldsymbol{\varphi} \ d\boldsymbol{\theta}$$

$$\stackrel{(1)}{\approx} \int_{\rho_i} p(x_i, \rho_i | z_i, \hat{\boldsymbol{\theta}}, \hat{\boldsymbol{\varphi}}, \alpha, \lambda) \ d\rho_i,$$

s.t. $(\hat{\boldsymbol{\theta}}, \hat{\boldsymbol{\varphi}}) = \arg\max_{\boldsymbol{\theta}, \boldsymbol{\varphi}} p(\boldsymbol{\theta}, \boldsymbol{\varphi} | \mathbf{z}_{-i}, \mathbf{x}_{-i}, \alpha, \lambda)$

$$= \int_{\rho_i} p(\rho_i | \hat{\boldsymbol{\varphi}}, z_i, \alpha) p(x_i | \rho_i, z_i, \hat{\boldsymbol{\theta}}, \lambda) \ d\rho_i$$

$$= \int_{\rho_i} p(\rho_i | \hat{\varphi}_{z_i}, \alpha) p(x_i | \rho_i, \hat{\theta}_{z_i}, \lambda) \ d\rho_i$$

$$\stackrel{(2)}{\approx} \quad \frac{\sum_{l=1}^{L} w_i \cdot p(x_i \mid \hat{\rho}_i^{\,l}, \hat{\theta}_{z_i}, \lambda)}{\sum_{l=1}^{L} w_i}$$

$$\text{s.t. } \{\hat{\rho}_i^{\,l}\}_{l=1}^{L} \sim q(\rho), \quad w_i = \frac{p(\hat{\rho}_i^{\,l} \mid \hat{\varphi}_{z_i}, \alpha)}{q(\hat{\rho}_i^{\,l})}$$

(1) approximates the posterior distribution of the parameters by its mode. The mode is computed using incremental hard-EM. Furthermore, the mode can be computed for each cluster's parameters independently. For every other data point $j$, perform an EM update:

$$\text{E:} \quad \hat{\rho}_j = \arg\max_{\rho_j} p(\rho_j \mid x_j, \hat{\theta}_{z_j}, \hat{\varphi}_{z_j})$$
$$= \arg\max_{\rho_j} p(\rho_j \mid \hat{\varphi}_{z_j}) p(x_j \mid \rho_j, \hat{\theta}_{z_j})$$

$$\text{M:} \quad \hat{\theta}_{z_j} = \arg\max_{\theta} p(\theta, \mid \{x_k, \rho_k \mid z_k = z_j\}, \lambda)$$
$$\hat{\varphi}_{z_j} = \arg\max_{\varphi} p(\varphi \mid \{\rho_k \mid z_k = z_j\}, \alpha)$$

(2) uses importance sampling in order to reduce the number of data transformations that need to be performed. Computing $p(x_i \mid \hat{\rho}_i^{\,l}, \hat{\theta}_{z_i}, \lambda)$ requires transforming the data point, which is the most computationally expensive single operation for this sampler. Thus it would be wise to reuse the samples, $\{\hat{\rho}_i^{\,l}\}_{l=1}^{L}$, across different clusters. We achieve this through importance sampling, which proceeds by sampling a set of transformation parameters from a proposal distribution, $q(\rho)$ and using those samples for all the clusters by reweighting them differently for each cluster. This is a large computational saving since the number of data transformation operations performed in a single iteration of this sampler is now independent of the number of clusters. Furthermore, the quality of approximation is controlled by the number of samples, $L$, generated.

To further increase the efficiency of the sampler, we approximate the maximization in the E-step by reusing the samples and selecting the one that maximizes $p(\rho_j \mid x_j, \hat{\theta}_{z_j}, \hat{\varphi}_{z_j})$. This avoids the direct maximization operation in the E-step which can be expensive. While not adopted in this work, further computational gains might be achieved at the expense of memory by storing and reusing samples (i.e. transformed data points) across iterations and reweighting them accordingly.

Thus our sampler iterates over every point in the data set, samples a cluster assignment and then updates $\hat{\theta}$ and $\hat{\varphi}$ for the sampled cluster. It also updates its own transformation parameter, $\hat{\rho}_i$ in the process.

**Summary.** We presented two samplers for our joint alignment and clustering model. Both samplers work well in practice, but the second is more efficient. For both samplers, every iteration begins by randomly permuting the order of the points and the DP concentration parameter is resampled using auxiliary variable methods [10]. As in the BA model, we cache the sufficient statistics for every cluster which can be updated efficiently as points are reassigned to clusters to allow for efficient likelihood and mode computation.

### 3.2 Incorporating Labelled Examples

The model presented in the previous section was used without any supervision. Supervision here refers to the ground-truth labels for some of the data points or the correct number of clusters. However, there are many scenarios where this information is available and would be advantageous to incorporate.

It is straightforward to modify the joint alignment and clustering model to accommodate such labelled examples. Lets assume we have positive examples for each cluster as well as a large data set of unlabeled examples. Before attempting to align and cluster the unlabeled examples, we would initialize several clusters and assign the positive examples to their respective clusters. By assigning these examples to their clusters and updating the sufficient statistics accordingly, the cluster parameters have incorporated the positive examples. Depending on the strength of the priors (i.e. the hyperparameters) and the number of positive examples per cluster it may be necessary to add the positive examples several times. The stronger the prior, the more times the positive examples need to be replicated. Note that replicating the positive examples does not increase memory usage since we only store the sufficient statistics for each cluster.

If the labelled portion contains positive examples for all the clusters, then setting the concentration parameter of the DP to 0 would prevent additional, potentially unnecessary, clusters from being created.

### 3.3 Experiment: Alignment and Clustering of Digits

We evaluated our unsupervised and semi-supervised models on two challenging data sets. The first contains 100 images of the digits "4" and "9", which are the two most similar and confusing digit classes (the performance of KMeans on this data set is close to random guessing). The second contains the 200 images of all 10 digit classes used by Liu *et al.* [24].[3] For the second data set we used the Histogram of Oriented Gradients (HOG) feature representation [7] used by Liu *et al.* to enable a fair comparison.

For both digit data sets we compared several algorithms using the same two metrics reported by Lui *et al.*: *alignment*

---
[3] Liu *et al.* also evaluated their model on 6 Caltech-256 categories and the CEAS face data set. For both data sets they randomly selected 20 images from each category. We found that the difficulty of a data set varied greatly from one sample to another, so we reached out to the authors. Unfortunately, they were only able to provide us with the digits data set which we do use. The digits data set was the most difficult of the three.

| Algorithm | Digits 4 and 9 (Fig 1) | | All 10 digits (Fig 7) | |
|---|---|---|---|---|
| | Alignment | Clustering | Alignment | Clustering |
| KMeans | 4.18 (1.57) ± 0.031 | 54.0% | 4.88 (1.61) ± 0.033 | 62.5% |
| Infinite mixture model [10] | 3.64 (1.34) ± 0.036 | 86.0%, 4 | 4.87 (1.64) ± 0.037 | 69.5%, 13 |
| Congealing [21] | 2.11 (0.93) ± 0.019 | 83.0% | 3.51 (1.34) ± 0.029 | 70.5% |
| TIC [11] | – | – | 6.00 (1.1) | 35.5% |
| Unsupervised SAC [24] | – | – | 3.80 (0.9) | 56.5% |
| Semi-supervised SAC [24] | – | – | not reported | 73.7% |
| Unsupervised JAC [§ 3.1] | **1.44 (0.69)** ± **0.014** | **94.0%**, 2 | **2.38 (1.12)** ± **0.027** | **87.0%**, 12 |
| Semi-supervised JAC [§ 3.2] | 1.58 (0.79) ± 0.016 | **94.0%** | 2.71 (1.25) ± 0.028 | 82.5% |

Table 1: Joint alignment and clustering of images. The left subtable refers to the first digit data set comprising 100 images containing the digits "4" and "9" (Figure 1), while the right subtable refers to the second data set comprising 200 images containing all 10 digits (the same data set used by Liu *et al*. [24], Figure 7). The alignment score columns contain three metrics that adhere to the following template: mean (standard deviation) ± standard error. The number following the clustering accuracy in the "Infinite mixture model" and "Unsupervised JAC" rows is the number of clusters that the model discovered (i.e. chose to represent the data with). On both data sets, our models significantly outperforms previous nonparametric alignment [21], joint alignment and clustering [11, 24], and nonparametric Bayesian clustering [10] models.

*score* measures the distance between pairs of aligned images assigned to the same cluster (we report the mean and standard deviation of all the distances, and the standard error[4]), and *clustering accuracy* is the Rand index with respect to the correct labels.

Table 1 summarizes the results on the models we evaluated:

- KMeans: we clustered the digits into the correct number of ground-truth classes (2 for the first data set, and 10 for the second) using the best of 200 KMeans runs.

- Infinite mixture model: removing the transformation/alignment component of our model reduces it to a standard Bayesian infinite mixture model. We ran this model to evaluate the advantage of joint alignment and clustering.

- Congealing: we ran congealing on all the images simultaneously and after alignment converged, clustered the aligned images using KMeans (with the correct number of ground-truth clusters). This allows us to evaluate the advantages of simultaneous alignment and clustering over alignment followed by clustering.

- TIC, USAC and SSAC results are listed exactly as reported by Liu *et al*.

- Unsupervised JAC refers to our full nonparametric Bayesian alignment and clustering model (§ 3.1).

- Semi-supervised JAC refers to the semi-supervised variant of our alignment and clustering model (§ 3.2). We used a *single* positive example for each digit and set the DP concentration parameter to 0.

[4]The standard error here is defined as the sample standard deviation divided by the square root of the number of pairs.

Note that the alignment scores for KMeans and the infinite mixture model are not relevant since no alignment takes place in either of these two algorithms. They are only included to offer a reference for the alignment score when the data is not transformed.

As the results show, our models outperform previous work with respect to both alignment and clustering quality. We make three observations about these results:

1. Our unsupervised model outperformed the unsupervised model of Liu *et al*. by $30.5\%$, and our semi-supervised model outperformed their semi-supervised model by $8.8\%$. This is in addition to the significant improvement in alignment quality.

2. Our unsupervised model improved upon the standard infinite mixture model in terms of alignment quality, clustering accuracy, and correctness of the discovered number of clusters.

3. The number of clusters discovered by our unsupervised model is quite accurate. For the first data set the model discovered the correct number of clusters (see Figure 1), and for the second it needed two additional clusters (see Figure 7).

These positive results validate our joint alignment and clustering models and associated learning schemes. Furthermore, it provides evidence for the advantage of solving both alignment and clustering problems simultaneously instead of independently.

### 3.4 Experiment: Alignment and Clustering of Curves

We now present joint alignment and clustering results on a challenging curve data set of ECG heart data [19] that is helpful in identifying the heart condition of patients. This data set contains 46 curves. 24 represent a normal heartbeat, and 22 represent an abnormal heartbeat. We ran both congealing and our nonparametric Bayesian joint alignment and clustering model. In both cases we excluded the non-linear scaling in amplitude transformation since the amplitudes of the curves are helpful in classifying whether the curve is normal or abnormal.

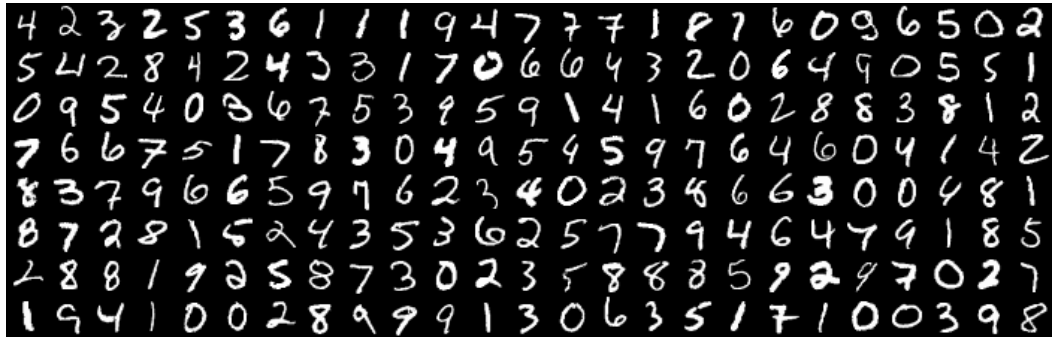
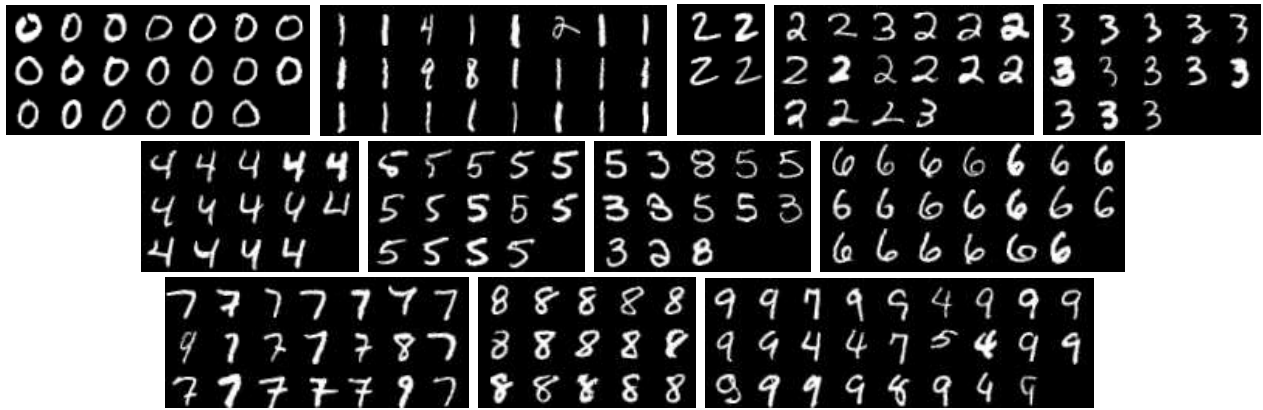

Figure 7: Unsupervised joint alignment and clustering of 200 images of all 10 digits. (Top) All 200 images provided to our model. (Bottom) The 12 clusters discovered and their alignments.

Our model discovered 5 clusters in the data set resulting in a clustering accuracy of $84.8\%$. Inspecting the clusters discovered by our model in Figure 8 highlights the fact that although the data set represents two groups (normal and abnormal), the curves do not naturally fall into two clusters and more are needed to explain the data appropriately. Figure 8 also displays the result of congealing the curves. Clustering the congealed curves into 2 clusters using the best of 200 KMeans runs results in a clustering accuracy of $71.7\%$. Clustering the congealed curves into 5 clusters in a similar manner results in a clustering accuracy of $76.1\%$.

The large improvement in clustering accuracy over congealing in addition to a much cleaner alignment result (Figure 8) highlights the importance of explicit clustering when presented with a complex data set. Furthermore it showcases our models ability to perform equally well on both image and curve data sets.

## 4 Discussions

In this section we discuss the adaptation of our joint alignment and clustering model to both online (when the data arrives at intervals) and distributed (when multiple processors are available) settings. Both of these adaptations are applicable to the unsupervised and semi-supervised settings.

### 4.1 Online Learning

There are several scenarios where online alignment and clustering may be helpful. Consider for instance a very large data set that cannot fit in memory or the case where the data set is not available up front but arrives over an extended period of time (such as in a tracking application).

An advantage of our model that has not yet been raised is its ability to easily adapt to an online setting where only a portion of the data set is available in the beginning. This is due to our use of conjugate priors and distributions in the exponential family which enable us to efficiently summarize an entire cluster through its sufficient statistics. Consequently, we can align/cluster the initial portion of the data set and save out the sufficient statistics for every cluster after each iteration (for both the data and transformations). Then as new data arrives, we can load in the sufficient statistics and use them to guide the alignment and clustering of the new

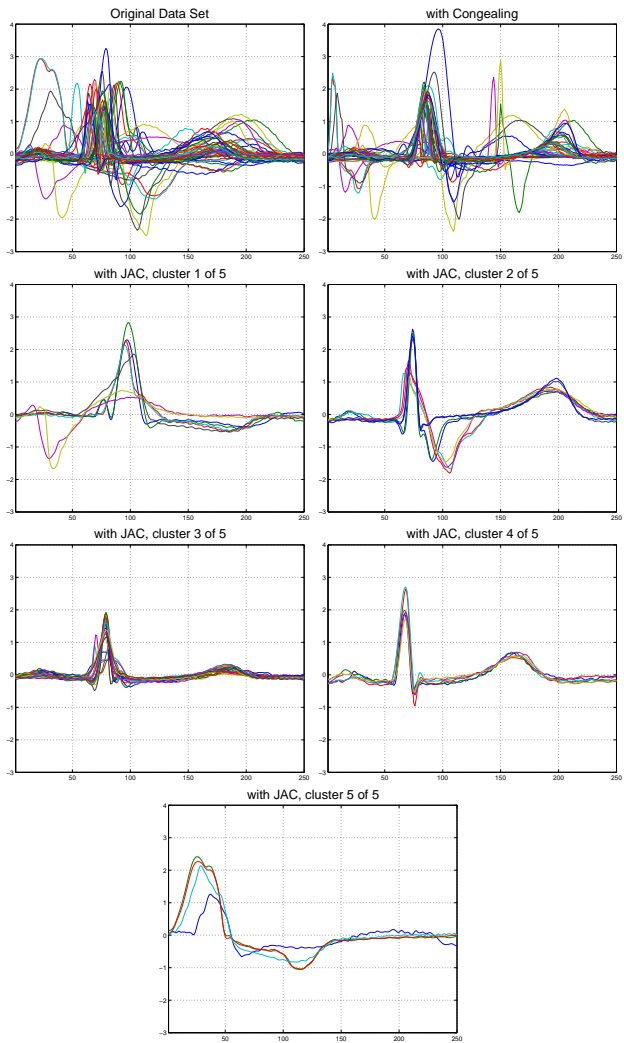

Figure 8: Joint alignment and clustering of ECG heart data. The first row displays the original data set (left) and the result of congealing (right). The last three rows display the 5 clusters discovered by our model.

data in lieu of the original data set which can now be discarded.

Given a sufficiently large initial data set, the alignment of a new data point using the procedure described above would be nearly identical to the result had that data point been included in the original set. This is true since the addition of a single point to an already large data set would have a negligible effect on the sufficient statistics. This process is also applicable to the Bayesian alignment model.

### 4.2 Distributed Learning

We now describe how to adapt our sampling scheme to a distributed setting using the MapReduce [8] framework. This facilitates scaling our model to large data sets in the presence of many processors. The key difference between the MapReduce implementation and the one described in § 3.1 is that the cluster parameters are updated once per sampling iteration instead of after each point's reassignment (i.e. using a standard sampler instead of a Rao-Blackwellized sampler).

A MapReduce framework involves two key steps, Map and Reduce. For our model the mapper would handle updating the transformation parameter and clustering assignment of a single data point, while the reducer would handle updating the parameters of a single cluster. More specifically, the input to each Map operation would be a data point along with a snapshot of the model parameters (the set of sufficient statistics that summarize the data set). The Map would output the updated cluster assignment and transformation parameter for that data point. The input to the Reduce step would then be all the data points that were assigned to a specific cluster (i.e. we would have a Reduce operation for every cluster created). The Reducer would then update the cluster parameters. Thus each sampling iteration is composed of a Map and Reduce stage.

## 5 Conclusion

We presented a nonparametric Bayesian joint alignment and clustering model that has been successfully applied to curve and image data sets. The model outperforms congealing and yields impressive gains in clustering accuracy over infinite mixture models. These results highlight the advantage of solving both alignment and clustering tasks simultaneously.

A strength of our model is the separation of the transformation function and sampling scheme, which makes it applicable to a wide range of data types, feature representations, and transformation functions. In this paper we presented results on three data types (2D points, 1D curves, and images), three transformation functions (point rotations, nonlinear curve transformations and affine image transformations), and two feature representations (identity and HOG).

In the future we foresee our model applied to a wide array of problems. Since curves are a natural representation for object boundaries [18], one of our goals is to apply our model to shape matching. We also intend to explore alternative parameter learning schemes based on variational inference [3, 20, 14].

### Acknowledgements

Special thanks to Gary Huang for many entertaining and helpful discussions. We would also like to thank Sameer Singh and the anonymous reviewers for helpful feedback. This work was supported in part by the National Science Foundation under CAREER award IIS-0546666 and Grant IIS-0916555.